\documentclass[journal]{EMO2007LBP}
\usepackage{algorithm}
\usepackage{algorithmic}
\usepackage{graphicx}

\newtheorem{definition}{Definition}[section]
\usepackage{url}       % url.sty is already installed on most LaTeX
\usepackage{amsmath}   % From the American Mathematical Society
\begin{document}

\title{Proposition of the Interactive Pareto Iterated Local Search Procedure -- Elements and Initial Experiments}

% author names
\author{Martin Josef Geiger%
\thanks{Martin Josef Geiger is with the Department of Industrial
Management (510A), University of Hohenheim, 70593 Stuttgart, Germany
(phone: 0049-711-45923462; fax: 0049-711-45923232; email:
mjgeiger@uni-hohenheim.de).}}
% note the % following the last name and also the first \thanks -
% instance, "A\textbf{} \textbf{}B" will typeset as "A B" not "AB". If you want
% "AB" then you have to do: "A\textbf{}\textbf{}B"

% *** Note that you probably will NOT want to include the author's name in ***
% *** the headers of peer review papers.                                   ***

% If you want to put a publisher's ID mark on the page
% (can leave text blank if you just want to see how the
% text height on the first page will be reduced by IEEE)
%\pubid{0000--0000/00\$00.00~\copyright~2002 IEEE}

%\specialpapernotice{(Invited Paper)}


\maketitle
\thispagestyle{empty}

\begin{abstract}
The article presents an approach to interactively solve
multi-objective optimization problems. While the identification of
efficient solutions is supported by computational intelligence
techniques on the basis of local search, the search is directed by
partial preference information obtained from the decision maker.

An application of the approach to biobjective portfolio
optimization, modeled as the well-known knapsack problem, is
reported, and experimental results are reported for benchmark
instances taken from the literature. In brief, we obtain encouraging
results that show the applicability of the approach to the described
problem.

%In order to stipulate a better understanding of the underlying
%structures of biobjective knapsack problems, we also study the
%characteristics of the search space of instances for which the
%optimal alternatives are known. As a result, optimal alternatives
%have been found to be relatively concentrated in alternative space,
%making the resolution of the studied instances possible with
%reasonable effort.

\end{abstract}



%=================================================================
%-----------------------------------------------------------------
% INTRODUCTION
%-----------------------------------------------------------------
%=================================================================
\section{\label{sec:introduction}Introduction}
\PARstart{A}{s} many problems of practical relevance are often
characterized by several criteria that simultaneously have to be
taken into consideration when solving the problem, multi-criteria
approaches play an increasingly important role in many application
areas of operations research, engineering, and computer science.
Approaches from the domain of multi criteria decision making
describe an alternative $x$, belonging to the set of feasible
alternatives $X$, by a set of objective functions $Z(x) = \left(
z_{1}(x), \ldots, z_{K}(x) \right)$. From a decision making
perspective, these functions describe attributes of the
alternatives, which are considered to be of relevance from the point
of view of a decision maker, in a quantitative way. As the aspects
and therefore the functions are however  often of conflicting
nature, not a single alternative exists being optimal for all
$z_{k}(x)$. Instead, a set of equally Pareto-optimal alternatives
can be found as introduced in the definitions below. Without loss of
generality, we assume in the following
Definitions~\ref{def:dominance} and \ref{def:efficiency} the
maximization of the components $z_{k}(x)$ of $Z(x)$.

\begin{definition}[Dominance]\label{def:dominance}
A vector $Z(x)$ is said to dominate $Z(x')$ iff $z_{k}(x) \geq
z_{k}(x') \forall k=1, \ldots, K \wedge \exists k \mid z_{k}(x) >
z_{k}(x')$. We denote the dominance of $Z(x)$ over $Z(x')$ with
$Z(x) \preceq Z(x')$.
\end{definition}

\begin{definition}[Efficiency,
Pareto-optimality]\label{def:efficiency} The vector $Z(x), x \in X$
is said to be efficient iff $\not\!\exists Z(x'), x' \in X \mid
Z(x') \preceq Z(x)$. The corresponding alternative $x$ is called
{\em Pareto-optimal}, the set of all Pareto-optimal alternatives
{\em Pareto-set $P$}.
\end{definition}

Two aspects play a vital role for solving multi-objective problems:

\begin{enumerate}
\item Search for optimal alternatives (the Pareto set $P$), supported by an
optimization approach. In comparison to single-objective
optimization approaches, the notion of optimality is here
generalized with respect to the set of simultaneously considered
optimality criteria.
\item Choice of a most-preferred solution by the decision maker of
the particular situation. While in single-objective optimization
problems, the choice of the most-preferred solution naturally
follows the identification of the (single) optimal solution, in
multi-objective problems an individual tradeoff between conflicting
criteria has to be resolved in a decision making procedure.
\end{enumerate}

Principally, three classes of combining search and decision making
exist \cite{horn:1997:incollection}:

\begin{enumerate}
\item {\em A priori} approaches reduce the multi-objective problem
into a single-objective problem by constructing a utility function
for the decision maker. The resolution of the problem then lies in
the identification of the solution which maximizes the chosen
utility function.
\item {\em A posteriori} approaches first identify the Pareto set
$P$ (or a close and representative approximation) and then resolve
the choice of a most-preferred solution within an interactive
decision making procedure.
\item {\em Interactive} approaches combine search and decision
making, presenting one or several solutions to the decision maker
and collecting preference information which is then used to further
guide the search for higher preferred alternatives.
\end{enumerate}

The question, which particular way to follow when solving
multi-objective optimization problems depends on several aspects. As
mentioned above, a priori approaches require comparably rich
preference information from the decision maker in order to reduce
the problem to a single-objective problem. Assuming this information
can be obtained, the resolution of the problem is straight-forward.
Unfortunately, this is not necessarily the case in many situations
as uncertainty about the precise appearance of the decision makers
utility function is often present. Also, the acceptance of a single
solution computed by a computational intelligence method may be a
problematic issue from a psychological point of view for some
decision makers as no further choice is possible.

A posteriori approaches on the other side do not require any
information from a decision maker. Instead, the Pareto set $P$ is
computed off-line, allowing the decision maker to perform other
tasks while waiting for the results of the optimization procedure.
Under the assumption that the set of optimality functions $Z(x)$ is
exhaustive, an important aspect when formulating multi-objective
models \cite{bouyssou:1990:incollection}, one element $x^{*} \in P$
can be identified in a later decision making procedure as the
most-preferred one. A potential drawback of this approach is
however, that the Pareto-set may be of large cardinality, resulting
in a necessary high computational effort when identifying the
Pareto-optimal alternatives. Also, many if not most $x \in P$ are
discarded in the decision making procedure as they do not meet the
individual, personal requirements of the decision maker.

Interactive approaches may overcome the problems of the above
described extreme ways of resolving multi-objective optimization
problem. On one hand, only partial preference information is
required when solving the problem as the search is guided in one
direction which may be changed when obtaining less-preferred
solutions. On the other hand, only fewer alternatives have to be
computed compared to the entire Pareto-set $P$. An important aspect
when implementing interactive approaches is, that the decision maker
needs to be present during the resolution procedure. Also, the
computation of the solutions has to be completed in little time as
the decision maker will have to wait for the results of the system.
With the increasing computational possibilities of modern,
affordable computer systems, the proposition and implementation
interactive approaches however becomes more and more attractive.

Recent approaches of computational intelligence techniques implement
interactive problem resolution procedures, e.\,g.\ on the basis of
Evolutionary Algorithms \cite{phelps:2003:article}, involving a
decision maker during search. While in these approaches the set of
criteria remains fixed during search, other concepts also include
the possibility of dynamically changing the relevant criteria when
searching for a most-preferred solution
\cite{geiger:2004:incollection}. Research in interactive
computational techniques is however a rather new field, and the
precise way of how to integrate articulated preferences in the
search process is still to be investigated in more detail.

In this article, we aim to contribute to the development of
interactive computational intelligence techniques for the resolution
of multi-objective optimization problems. While the search for
Pareto-optimal alternatives is done by metaheuristics on the basis
of local search, individual preferences guide the search in a
particular direction with the goal of identifying a subset of $P$
that is considered to be of interest to the decision maker. While
the idea is generic, it is tested on a particular application.

The article is organized as follows. In the following
Section~\ref{sec:problem:description}, the biobjective portfolio
optimization problem is introduced and a quantitative optimization
model is presented. We also briefly review existing approaches from
the literature that have been used to solve this problem. %In order
%to obtain a better understanding of the underlying structures of the
%particular problem, an analysis of the search space has been carried
%out which follows in Section~\ref{sec:search:space}.
An interactive procedure to solve the problem is proposed in
Section~\ref{sec:solution:approach}. Experimental investigations on
benchmark instances taken from literature follow in
Section~\ref{sec:experiments}, and conclusions are drawn in
Section~\ref{sec:conclusions}.

%=================================================================
%-----------------------------------------------------------------
% PROBLEM DESCRIPTION
%-----------------------------------------------------------------
%=================================================================
\section{\label{sec:problem:description}Problem description}
The multi-objective portfolio optimization problem consists in
selecting a subset of assets from a set of $n$ possible investment
possibilities such that several criteria, mainly the profit and risk
of the resulting portfolio, are optimized. More formally, the
following program has to be solved.

\begin{equation}
\max z_{k}(x) = \sum_{j=1}^{n} p_{j}^{k} \mathrm{x}_{j} \quad
\forall k = 1, \ldots, K
\end{equation}

s.\,t.\

\begin{equation}\label{eqn:weight}
\sum_{j=1}^{n} c_{j} \mathrm{x}_{j} \leq C
\end{equation}

\begin{equation}\label{eqn:binary}
\mathrm{x}_{j} \in \{ 0, 1 \}
\end{equation}

Each alternative $x$ consists of an $n$-dimensional decision vector
$x = ( \mathrm{x}_{1}, \ldots, \mathrm{x}_{n} )$ which defines for
each asset $j$ whether it is included in the portfolio
($\mathrm{x}_{j} = 1$) or not ($\mathrm{x}_{j} = 0$), given in
constraint~(\ref{eqn:binary}). Side constraint~(\ref{eqn:weight})
ensures that a given capacity (e.\,g.\ a budget) of $C$ is not
exceeded, and typically the nonnegative coefficients $c_{j}$ relate
to each other as $c_{j} \leq C \,\, \forall j=1, \ldots, n$ and
$\sum_{j=1}^{n} c_{j} > C$. The nonnegative coefficients $p_{j}^{k}$
express for each $\mathrm{x}_{j}$ the contribution (e.\,g.\ the
profit) of asset $j$ to objective $k$.

As the objectives conflict with each other, a set of Pareto-optimal
alternatives $P$ exists among which the choice of a most-preferred
solution $x^{*} \in P$ has to be made.

Unfortunately, the problem, also referred to as the so called
knapsack problem, is $\mathcal{NP}$-hard \cite{garey:1979:book},
even for a $K = 1$. For single-objective problems, relatively good
branch-and-bound algorithms have been developed
\cite{martello:1990:book}, other approaches are based on dynamic
programming \cite{klamroth:2000:article}. The multi-objective case
with $K \geq 2$ has been tackled again with branch-and-bound
\cite{ulungu:1997:incollection} and several heuristics, mainly local
search as e.\,g.\ Simulated Annealing \cite{ulungu:1999:article}, or
Tabu Search \cite{gandibleux:2000:article}.

\section{\label{sec:solution:approach}Solution approach}
Based on the discussion of different problem resolution strategies
in Section~\ref{sec:introduction} %and the insight in the structure
%of biobjective knapsack problems gained in
%Section~\ref{sec:search:space},
we propose an interactive metaheuristic that brings search and
decision making together in a combined approach.

\begin{figure}[!ht]
\begin{center}
\includegraphics[width=8cm]{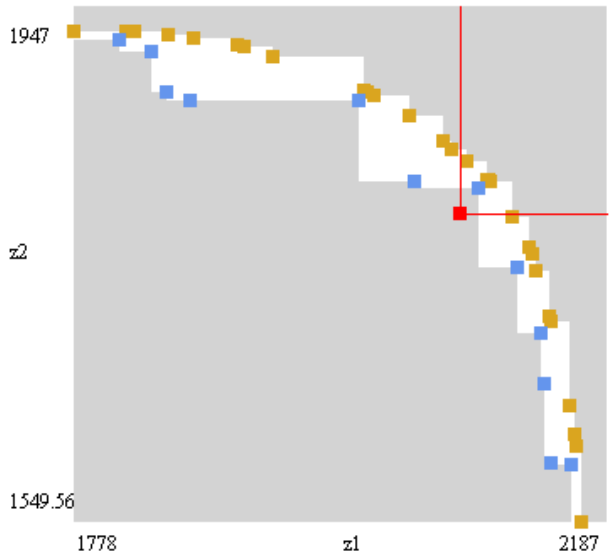}
\caption{\label{fig:screenshot}Lower/ upper approximation and
determination of the goal vector}
\end{center}
\end{figure}

In the first step of the problem resolution approach, the decision
maker is provided with lower and upper bounds of the particular
problem. For the knapsack problem, this can be done by a set of
weights $W$ and relaxing the binary constraint~\ref{eqn:binary} for
the computation of upper bounds. An example of a visual output is
given in Figure~\ref{fig:screenshot}, plotting the lower bound sets
in blue, the upper bound sets in orange.

By means of the visualization of the lower and upper bounds, the
decision maker may easily see the area in which Pareto-optimal
solutions can be found. It becomes equally clear what values of
$Z(x)$ are not possible as they would dominate an upper bound,
enabling the decision maker to develop realistic expectations of a
most-preferred solution $x^{*}$. In other words, objective values
which are not achievable, as they lie outside the boundaries of the
upper bounds, can be excluded from the decision making procedure.

The search for Pareto-optimal solutions is then guided by the
articulation of a reference point $R = (r_{1}, \ldots, r_{K})$,
given in red color in Figure~\ref{fig:screenshot}. This reference
point defines a cone in outcome space which is used to specify an
area in which the most-preferred solution $x^{*}$ is expected. After
having successfully implemented an a posteriori strategy based on
the same principle in which the identification of a most-preferred
solution is supported by the progressive articulation of aspiration
levels \cite{geiger:2006:moscheduling:system}, we use this idea here
to interactively guide the search.

\begin{algorithm}[!ht]%
\caption{\label{alg:optimization:shellpils}Optimization framework}
\begin{algorithmic}[1]%
\STATE{Compute a first approximation $P^{approx}$ of $P$ using a set of weights $W$.}%
\STATE{Present $P^{approx}$ to the decision maker}%
\STATE{Obtain a reference point $R$ from the decision maker} %
\REPEAT%
    \STATE{Compute $P_{R}^{approx}$}
    \WHILE{$R$ has not been changed {\bf or} termination criterion has not been met}
        \STATE{Search for Pareto-optimal alternatives in the cone defined by $R$ by means of a pre-defined local search metaheuristic}%%
        \STATE{Constantly update $P_{R}^{approx}$ while searching for Pareto-optimal solutions}%%
        \STATE{Constantly update the visualization of $P_{R}^{approx}$, showing the results to the decision maker}%%
    \ENDWHILE
\UNTIL{termination criterion is met}
\end{algorithmic}%
\end{algorithm}%

\begin{figure}[!ht]
\begin{center}
\includegraphics{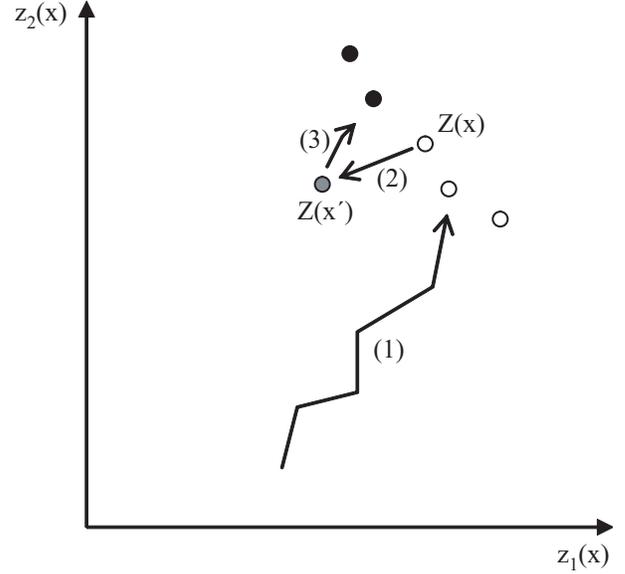}
\caption{\label{fig:pils}Pareto Iterated Local Search}
\end{center}
\end{figure}

The cone defined by $R$ is used to compute the set of alternatives
$P_{R}^{approx}$ that dominate the reference point and therefore lie
in the interior of the cone. In brief, $P_{R}^{approx}$ contains all
elements $x \in P^{approx} \mid z_{k}(x) \geq r_{k} \forall k=1,
\ldots, K$, therefore $P_{R}^{approx} \subseteq P^{approx}$. These
alternatives are used by the Pareto Iterated Local Search
metaheuristic, whose principle is sketched in Figure~\ref{fig:pils}
and discussed in the following. Search continues until the decision
maker terminates the process. This is going to be the case when a
solution $x^{*} \in P_{R}^{approx}$ is found which meets the
individual requirements and expectations of the decision maker as
close as possible.

Starting from an initial solution, a local search run is performed
using a neighborhood operator until no further improvement is
possible with simple local modifications. During this step, an
archive of alternatives is maintained which contains all
non-dominated alternatives found during search. While in the general
procedure of Pareto Iterated Local Search
\cite{geiger:2006:pils:scheduling} all non-dominated alternatives
are kept, we here restrict the procedure to keep only alternatives
in the cone defined by $R$ for further modifications/ improvements.
In Figure~\ref{fig:pils}, this stage of the procedure is visualized
as step (1), with the results shown as white points in outcome
space.

After having obtained a set of locally optimal alternatives, one of
them is picked at random, see $Z(x)$ in Figure~\ref{fig:pils},
perturbed into another alternative $x'$ using some other
neighborhood (2), and search is continued from here (3). As it can
be seen in Figure~\ref{fig:pils}, the perturbed solution may be
dominated by one or several elements of $P_{R}^{approx}$, which has
to be accepted when overcoming local optimality. In result, the
metaheuristic iterates in interesting areas of the search space as
opposed to restarting search from some other solution. This
principle, known from Iterated Local Search
\cite{lourenco:2003:incollection}, has been already successfully
applied to other problems in which considerable
fitness-distance-correlations have been found
\cite{boese:1996:phdthesis}.

The computations of the metaheuristic are continued, constantly
updating the plot of the Pareto-optimal alternatives in outcome
space. During the problem resolution procedure, the decision maker
is allowed modify the reference point, shifting the focus of the
computations towards other regions. The problem resolution procedure
terminates with the identification of a most-preferred solution
$x^{*}$.

%=================================================================
%-----------------------------------------------------------------
% EXPERIMENTS
%-----------------------------------------------------------------
%=================================================================
\section{\label{sec:experiments}Experimental investigations}
\subsection{Experimental setup}
The interactive Pareto Iterated Local Search has been tested on a
benchmark instance taken from \cite{gandibleux:2000:article}. Apart
from the fact that the optimal alternatives of these instances are
known as mentioned above, the data sets are widely used for
experimental investigations and comparison. In choosing them, we
hope to provide a basis for fair and representative comparison. The
data of the instance can be obtained from the internet homepage of
the International Society on Multiple Criteria Decision Making under
\url{http://www.terry.uga.edu/mcdm/}.

Local modifications of alternatives $x$ are done by randomly picking
a single decision variable $\mathrm{x}_{j} \mid \mathrm{x}_{j} = 1$,
changing its value to $\mathrm{x}_{j} = 0$, and randomly changing
the value of other randomly chosen decision variables to 1 until no
additional asset may be added to the solution.

We applied this local search neighborhood to each element in
$P_{R}^{approx}$ until a dominating alternative has been found,
replacing the alternative, or a subsequent number of 100
unsuccessful iterations has been tested on each element in
$P_{R}^{approx}$. Then, the perturbation is applied to a randomly
picked element in $P_{R}^{approx}$. The alternative is perturbed by
changing two randomly chosen variables $\mathrm{x}_{j} =
\mathrm{x}_{l} = 1$ to 0 and refilling up the knapsack to the
capacity $C$ by randomly selected other assets. In this sense, the
perturbation is similar to the regular neighborhood, only that more
decision variables are involved, leading to a bigger jump in the
search space while keeping most of the characteristics of the
perturbed alternative at the same time. The search then continues
from the alternative which has been obtained through perturbation as
described in Section~\ref{sec:solution:approach}.

In order to simulate the individual preference articulation of the
decision maker, three reference points have been defined as given in
table~\ref{tbl:goal:vectors}, one in the \lq{}knee-region\rq{}
\cite{rachmawati:2006:inproceedings} and two in the extreme areas of
either one of the objective functions.

\begin{table}[!ht]
\begin{center}
%% increase table row spacing, adjust to taste
\renewcommand{\arraystretch}{1.3}
\caption{Reference points} \label{tbl:goal:vectors}
\begin{tabular}{lll}
\hline
Model & Reference point & Vector\\
\hline%%
%2KP50-11 & ref. \#1 & $r_{1} = 395, r_{2} = 550$\\
%2KP50-11 & ref. \#2 & $r_{1} = 531, r_{2} = 428$\\
%2KP50-11 & ref. \#3 & $r_{1} = 616, r_{2} = 354$\\
2KP50-50 & ref. \#1 & $r_{1} = 1807, r_{2} = 1924$\\
2KP50-50 & ref. \#2 & $r_{1} = 2094, r_{2} = 1800$\\
2KP50-50 & ref. \#3 & $r_{1} = 2166, r_{2} = 1574$\\
\hline
\end{tabular}
\end{center}
\end{table}

100 test runs have been carried out with each reference point,
keeping the approximations $P_{R}^{approx}$ for further analysis. In
each test run 100,000 iterations have been allowed before
terminating the search.

The quality of the computed approximations has been analyzed using
the $M$ metric, given in expression~\ref{eqn:M}. $M$ measures the
percentage of identified Pareto-optimal alternatives in the cone
defined by $R$.

\begin{equation}\label{eqn:M}
M = \frac{|P_{R}^{approx}\cap P_{R}|}{|P_{R}|}
\end{equation}

\subsection{Results}
\begin{figure}[!ht]
\begin{center}
\includegraphics[width=8.5cm]{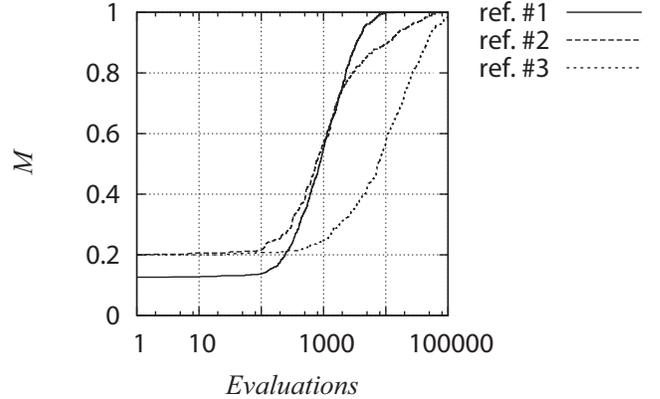}
\caption{\label{fig:results:50-50}Results for 2KP50-50}
\end{center}
\end{figure}

Based on the data gathered in the experiments, the arithmetic mean
values of $M$ have been computed, depending number of evaluations of
the metaheuristic. These average values, given in
Figure~\ref{fig:results:50-50}, clearly show that the iPILS
metaheuristic successfully identified the Pareto-optimal
alternatives in the particular areas of the reference points.
However, there does not turn out to be a consistent difference for
the three chosen reference points within the same instance.

%\begin{figure}[!ht]
%\begin{center}
%\includegraphics[width=8.5cm]{fig_2KP50-11.eps}
%\caption{\label{fig:results:50-11}Results for 2KP50-11}
%\end{center}
%\end{figure}

While some approximations $P_{R}^{approx}$ already contain a
Pareto-optimal alternative right from the start, no overall
advantage for the final approximation quality results from this
circumstance. Some areas are faster approximated, e.\,g.\ the one of
reference point \#1, others take considerable more time, see
reference point \#3.

%=================================================================
%-----------------------------------------------------------------
% CONCLUSIONS
%-----------------------------------------------------------------
%=================================================================
\section{\label{sec:conclusions}Conclusions}
The article presented an interactive method for the resolution of
multi-objective optimization problems. The concept is based on the
articulation of a reference point which expresses, from the point of
view of the decision maker, an interesting area in outcome space.
The computation of Pareto-optimal solutions is consequently focused
on that region, identifying optimal solutions. In order to overcome
locally optimal alternatives and converge to the front of the
efficient solutions, a metaheuristic based on Iterated Local Search
has been implemented.

Tests on the biobjective portfolio optimization problem have been
carried out. In order to simulate a human decision maker, we assumed
three different reference points for the investigated problem
instance. Each one was chosen with respect to lower and upper bound
sets that serve as an orientation for the decision maker. The
problem resolution technique successfully solved the investigated
benchmark instance, independent from the particular reference point.

Based on the investigations and experiments carried out, we conclude
that the proposed concept may be a useful tool for solving
multi-objective optimization problems, given the possibility to
appropriately compute lower and upper bound sets of the particular
problem. The results are encouraging, and deeper investigations on
more instances will follow to support the results of the study.

\section*{Acknowledgment}
The participation in the 4th International Conference on
Evolutionary Multi-Criterion Optimization (EMO2007) has been
partially supported by the Deutsche Forschungsgemeinschaft (DFG),
grant no.\ 535530.

%=================================================================
%-----------------------------------------------------------------
% BIBLIOGRAPHY
%-----------------------------------------------------------------
%=================================================================

\bibliography{../../../lit_bank,../../../lit_bank_nv,../../../lit_bank_datei}

\begin{thebibliography}{10}
\providecommand{\url}[1]{#1}
\csname url@rmstyle\endcsname
\providecommand{\newblock}{\relax}
\providecommand{\bibinfo}[2]{#2}
\providecommand\BIBentrySTDinterwordspacing{\spaceskip=0pt\relax}
\providecommand\BIBentryALTinterwordstretchfactor{4}
\providecommand\BIBentryALTinterwordspacing{\spaceskip=\fontdimen2\font plus
\BIBentryALTinterwordstretchfactor\fontdimen3\font minus
  \fontdimen4\font\relax}
\providecommand\BIBforeignlanguage[2]{{%
\expandafter\ifx\csname l@#1\endcsname\relax
\typeout{** WARNING: IEEEtran.bst: No hyphenation pattern has been}%
\typeout{** loaded for the language `#1'. Using the pattern for}%
\typeout{** the default language instead.}%
\else
\language=\csname l@#1\endcsname
\fi
#2}}

\bibitem{horn:1997:incollection}
J.~Horn, ``Multicriterion decision making,'' in \emph{Handbook of Evolutionary
  Computation}, T.~B\"{a}ck, D.~B. Fogel, and Z.~Michalewicz, Eds.\hskip 1em
  plus 0.5em minus 0.4em\relax Bristol: Institute of Physics Publishing, 1997,
  ch. F1.9, pp. F1.9:1--F1.9:15.

\bibitem{bouyssou:1990:incollection}
D.~Bouyssou, ``Building criteria: A prerequisite for {MCDA},'' in
  \emph{Readings in Multiple Criteria Decision Aid}, C.~{Bana E Costa},
  Ed.\hskip 1em plus 0.5em minus 0.4em\relax Heidelberg: Springer Verlag, 1990,
  pp. 58--80.

\bibitem{phelps:2003:article}
S.~P. Phelps and M.~K\"{o}ksalan, ``An interactive evolutionary metaheuristic
  for multiobjective combinatorial optimization,'' \emph{Management Science},
  vol.~49, pp. 1726--1738, 2003.

\bibitem{geiger:2004:incollection}
M.~J. Geiger and S.~Petrovic, ``An interactive multicriteria optimisation
  approach for scheduling,'' in \emph{Artificial Intelligence Applications and
  Innovations}, M.~Bramer and V.~Devedzic, Eds.\hskip 1em plus 0.5em minus
  0.4em\relax Boston, Dordrecht, London: Kluwer Academic Publishers, 2004, pp.
  475--484.

\bibitem{garey:1979:book}
M.~R. Garey and D.~S. Johnson, \emph{Computers and Intractability---A Guide to
  the Theory of $NP$-Completeness}.\hskip 1em plus 0.5em minus 0.4em\relax San
  Francisco, CA: W. H. Freeman and Company, 1979.

\bibitem{martello:1990:book}
S.~Martello and P.~Toth, \emph{Knapsack Problems: Algorithms and Computer
  Implementations}.\hskip 1em plus 0.5em minus 0.4em\relax Chichester, New
  York, Brisbane, Toronto, Singapore: John Wiley \& Sons, 1990.

\bibitem{klamroth:2000:article}
K.~Klamroth and M.~M. Wiecek, ``Dynamic programming approaches to the multiple
  criteria knapsack problem,'' \emph{Naval Research Logistics}, vol.~47, pp.
  57--76, 2000.

\newpage 

\bibitem{ulungu:1997:incollection}
E.~L. Ulungu and J.~Teghem, ``Solving multi-objective knapsack problem by a
  branch-and-bound procedure,'' in \emph{Multicriteria Analysis}, J.~N.
  Climaco, Ed.\hskip 1em plus 0.5em minus 0.4em\relax Berlin, Heidelberg, New
  York: Springer Verlag, 1997, pp. 269--278.

\bibitem{ulungu:1999:article}
E.~L. Ulungu, J.~Teghem, P.~H. Fortemps, and D.~Tuyttens, ``{MOSA} method: A
  tool for solving multiobjective combinatorial optimization problems,''
  \emph{Journal of Multi-Criteria Decision Making}, vol.~8, pp. 221--236, 1999.

\bibitem{gandibleux:2000:article}
X.~Gandibleux and A.~Freville, ``Tabu search based procedure for solving the
  0-1 multiobjective knapsack problem: the two objectives case,'' \emph{Journal
  of Heuristics}, vol.~6, no.~3, pp. 361--383, 2000.

\bibitem{geiger:2006:moscheduling:system}
M.~J. Geiger, ``Solving multi-objective scheduling problems---an integrated
  systems approach,'' in \emph{Artificial Intelligence in Theory and Practice},
  ser. IFIP International Federation for Information Processing, M.~Bramer,
  Ed.\hskip 1em plus 0.5em minus 0.4em\relax New York: Springer Verlag, 2006,
  vol. 217, pp. 493--502, ISBN 0-387-34654-6.

\bibitem{geiger:2006:pils:scheduling}
M.~J. Geiger, ``The {PILS} metaheuristic and its application to multi-objective
  machine scheduling,'' in \emph{Multicriteria Decision Making and Fuzzy
  Systems -- Theory, Methods and Applications}, ser. Industrial and applied
  mathematics, K.-H. K\"{u}fer, H.~Rommelfanger, C.~Tammer, and K.~Winkler,
  Eds.\hskip 1em plus 0.5em minus 0.4em\relax Aachen: Shaker Verlag, 2006, pp.
  43--58, ISBN 3-8322-5540-0.

\bibitem{lourenco:2003:incollection}
H.~R. Louren\c{c}o, O.~Martin, and T.~St\"{u}tzle, ``Iterated local search,''
  in \emph{Handbook of Metaheuristics}, ser. International Series in Operations
  Research \& Management Science, F.~Glover and G.~A. Kochenberger, Eds.\hskip
  1em plus 0.5em minus 0.4em\relax Boston, Dordrecht, London: Kluwer Academic
  Publishers, 2003, vol.~57, ch.~11, pp. 321--353.

\bibitem{boese:1996:phdthesis}
K.~D. Boese, ``Models for iterative global optimization,'' Ph.D. dissertation,
  University of California at Los Angeles, Los Angeles, California, 1996.

\bibitem{rachmawati:2006:inproceedings}
L.~Rachmawati and D.~Srinivas, ``A multi-objective genetic algorithm with
  controllable convergence on knee regions,'' in \emph{2006 IEEE Congress on
  Evolutionary Computation}, Sheraton Vancouver Wall Centre Hotel, Vancouver,
  BC, Kanada, Juli 2006, pp. 6807--6814, ISBN 0-7803-9489-5.

\end{thebibliography}
\bibliographystyle{IEEEtran.bst}

\end{document}